\documentclass[runningheads]{llncs}

\usepackage{epsfig}

\usepackage{xspace}

\usepackage{amssymb}
\usepackage{longtable}

\usepackage{color}

\begin{document}

\pagestyle{headings}

\mainmatter

\title{On Laughter and Speech-Laugh, Based on Observations of Child-Robot Interaction}
\titlerunning{Laughter in Child-Robot Interaction}

  \author{Anton Batliner\inst{1}
  \and Stefan Steidl\inst{1}
  \and Florian Eyben\inst{2}
   \and Bj{\"o}rn Schuller\inst{2} 
  }

  \authorrunning{Anton Batliner et al.}

  \institute{ Pattern Recognition Lab, Department of Computer Science,  Friedrich-Alexander-University Erlangen-Nuremberg (FAU), Martensstr. 3, 91058 Erlangen, Germany  \\
  \email{\{batliner,steidl\}@informatik.uni-erlangen.de}
   \\
  \texttt{http://www5.informatik.uni-erlangen.de}
  \and
  Institute for Human-Machine Communication,
  Technische Universit{\"a}t M{\"u}nchen (TUM), Germany \\
  \email{\{eyben,schuller\}@tum.de}
  }

\maketitle

\begin{abstract}
 In this article, we study laughter found in child-robot interaction where it had not been prompted intentionally. Different types of laughter and speech-laugh are annotated and processed. In a descriptive part, we report on  the position of laughter and speech-laugh in syntax and dialogue structure, and on communicative functions.  In a second part, we report on automatic classification performance and on acoustic characteristics, based on extensive feature selection procedures.
 
\end{abstract}

\def\eg{e.\,g.\xspace}
\def\ie{i.\,e.\xspace}
\def\wrt{w.\,r.\,t.\xspace}

\pagenumbering{arabic}

\section{Introduction}
\label{Introduction}

Until the mid nineties, automatic speech recognition (ASR) concentrated on word recognition and subsequently, on processing of higher linguistic information such as dialogue acts, semantic saliency, recognition of accents and boundaries,  etc. Paralinguistic information was  normally not accounted for and treated the same way as non-linguistic events such as technical noise. Paralinguistic  information  is either modulated onto the speech chain as, \eg voice quality such as laryngealisations \cite{Batliner07-LAE}, or it is interspersed between the words, such as filled pauses or laughter. More generally, laughter belongs to the group of the so-called affect bursts \cite{Schroeder00-ESO} which are partly words -- including specific semantics -- partly non-linguistic events. 

Normally, laughter is conceived of as a non-linguistic or paralinguistic event.
As one possibility to express emotions (especially joy), it has been dealt with already by Darwin \cite{Darwin72-TEO}; studies on its acoustics, however, as well as its position in linguistic context -- in the literal meaning of the word (where it can be found in the word chain, cf.\ \cite{Provine93-LPS}), and in the figurative sense (status and function) -- started more or less at the same time as ASR started to deal with paralinguistic phenomena. The acoustics of laughter are for example described in \cite{Bacharowski01-TAF,Trouvain01-PAO} and in further studies referred to in these articles. In \cite{Trouvain03-SPU}  an overview of phenomena and terminology is given. The context of laughter is addressed in \cite{Campbell05-NLM,Campbell07-WWL} (different types of laughter and their function, different addressees in communication), and in \cite{Laskowski07-AOT} (distribution of laughter within multi-party conversations).  As for the automatic classification of laughter, cf.\ \cite{Truong05-ADO,Truong07-ADB,Petridis08-ALD} and further studies referred to in these articles. 

 The scenario of the present study is child-robot interaction. Laughter is not elicited intentionally; the children had to accomplish different tasks by giving the robot -- Sony's dog-like pet-robot Aibo -- commands. As far as we can see, this specific combination `children\,+\,pet-robot' has not yet been addressed in studies on laughter so far. 
%

\section{Overview}
\label{Overview}

In this article, we deal with laughter as an event on the time axis which can be delimited (segmented) the same way as words can, and with speech-laugh, 
\ie laughter modulated onto speech, which is co-extensive with the word it `belongs to'. 
Throughout, we will use small capitals (\textsc{speech-laugh}, \textsc{laughter}) -- but italics (\textit{SL} vs.\ \textit{L}) if abbreviated -- when referring to the phenomena that have been  annotated and processed. When we refer to the generic term, we simply use the regular font (`laughter').
After the presentation of the database in Section \ref{Database}, we describe  the annotation of emotional user states, of the different types of laughter established, and of syntactic boundaries in Section \ref{Annotation}. Section \ref{Results} reports on empirical findings: the duration of laughter,  
its syntactic position, its communicative function, and  its position in the dialogue. 
Our interest in automatic processing, which is dealt with in Section \ref{Automatic-classification},  is twofold: of course, we are interested in classification performance. Moreover, we want to use the feature selection method employed to find out which acoustic characteristics can be found for laughter in general, and for the different subtypes of \textsc{speech-laugh} and \textsc{laughter} in particular.

\section{The Database}
\label{Database}

The database used is a German corpus of children communicating with
Sony's pet robot Aibo, the \emph{FAU Aibo Emotion Corpus}, cf.\ \cite{Batliner08-PEV,Steidl09-ACO,Schuller09-TI2}.
%
It can be considered as a corpus of spontaneous speech, because the
children were not told to use specific instructions but to talk to the
Aibo as they would talk to a friend. Emotional, affective states
conveyed in this speech are not elicited explicitly (prompted) but
produced by the children in the course of their interaction with the
Aibo; thus they are fully naturalistic.
The children were led to believe that the Aibo was responding to their
commands, whereas the robot was actually controlled by a human
operator (Wizard-of-Oz, WoZ) using the `Aibo Navigator' software over
a wireless LAN (the existing Aibo speech recognition module was not
used).  The WoZ caused the Aibo to perform a fixed, predetermined
sequence of actions; sometimes the Aibo behaved disobediently, thus
provoking emotional reactions.  The data were collected at two
different schools from 51 children (age 10--13, 21 male, 30 female).
Speech was transmitted via a wireless head set (UT 14/20 TP SHURE
UHF-series with microphone WH20TQG) and recorded with a DAT-recorder
(sampling rate 48\,kHz, quantisation 16\,bit, down-sampled to
16\,kHz).  Each recording session took some 30 minutes. Because of the
experimental setup, these recordings contain a huge amount of silence
(reaction time of the Aibo), which caused a noticeable reduction of
recorded speech after raw segmentation; eventually we obtained almost
nine hours of speech.
The audio-stream was segmented automatically with a pause threshold of 1\,sec. into so-called  \emph{turns}.

 In planning the sequence of Aibo's actions, we tried to find a
good compromise between obedient and disobedient behaviour: we wanted
to provoke the children in order to elicit emotional behaviour, but of
course we did not want to run the risk that they break off the
experiment.
The children believed that the Aibo was reacting to their orders --
albeit often not immediately. In reality, the scenario was the
opposite: the Aibo always strictly followed the same screen-plot, and
the children had to align their orders to its actions.
By these means, it was possible to examine different children's
reactions to the very same sequence of Aibo's actions.
In the so-called `parcours' task, the children had to direct the Aibo
from START to GOAL; on the way, the Aibo had to fulfil some tasks and
had to sit down in front of three cups. This constituted the longest sub-task. 
In each of the other five tasks of the experiment, 
 the children were instructed to direct the Aibo towards one of
 several cups standing on the carpet. One of these cups was `poisoned'
 and had to be avoided.  The children applied different strategies to
 direct the Aibo. Again, all actions of Aibo were pre-determined.
 In the first task, Aibo was `obedient' in order to make the children
 believe that it would understand their commands. In the other
 tasks, Aibo was `disobedient'. In some tasks Aibo went directly
 towards the `poisoned' cup in order to evoke emotional speech from
 the children.
%
%
 No child broke off the experiment, although it could be clearly seen
 towards the end that some of them were bored and wanted to put an end
 to the experiment -- a reaction that we wanted to provoke.
 Interestingly, in a post-experimental questionnaire, all children
 reported that they had much fun and liked it very much; thus we can expect at least some instances of laughter indicating joy. At least two  different conceptualisations could be observed: in the first, the  Aibo was treated as a sort of remote-control toy (commands like
 \textit{``turn left'', ``straight on'', ``to the right''}); in the second, the
 Aibo was addressed the same way as a pet dog (commands like
 \textit{``Little Aibo doggy, now please turn left -- well done, great!''}
 or \textit{``Get up, you stupid tin box!''}), cf.\ \cite{Batliner08-PEV}.

Detailed information on the database is given in \cite{Steidl09-ACO}.\footnote{The book can be downloaded  from the web: 
\textit{http://www5.informatik.uni-erlangen.de/Forschung/Publikationen/2009/Steidl09-ACO.pdf}.}


\section{Annotation}
\label{Annotation}

\subsection{Emotion}
\label{Emotion}
Five labellers (advanced students of linguistics, 4 females, 1 male)
listened to the speech files in sequential order and annotated
independently from each other each word as neutral (default) or as
belonging to one of ten other classes, which were obtained by 
inspection of the data. This procedure was iterative and supervised by
an expert.
The sequential order of labelling does not distort the linguistic and
paralinguistic message.  Needless to say, we do not claim that these
classes represent children's emotions (emotion-related user states) in
general, only that they are adequate for the modelling of these
children's behaviour in this specific scenario.
We resort to majority voting (henceforth MV): if three or more
labellers agree, the label is attributed to the word; if four or five
labellers agree, we assume some sort of prototypes.  The following raw
labels were used; in parentheses, the number of cases with MV is
given: \textit{joyful} (101), \textit{surprised} (0),
\textit{emphatic} (2528), \textit{helpless} (3), \textit{touchy}, \ie,
irritated (225), \textit{angry} (84), \textit{motherese} (1260),
\textit{bored} (11), \textit{reprimanding} (310), \textit{rest}, \ie
non-neutral, but not belonging to the other categories (3),
\textit{neutral} (39169); 4707 words had no MV; all in all, there were
48401 words.  \textit{joyful} and \textit{angry} belong to the `big', basic 
emotions \cite{Ekman99-BE}, the other ones rather to `emotion-related/emotion-prone'
user states but have been listed in more extensive catalogues of
emotion/emotion-related terms, \eg `reproach' (\ie
\textit{reprimanding}), \textit{bored}, or \textit{surprised} in
\cite{Ortony88-TCS}.  The state \textit{emphatic} has been introduced
because it can be seen as a possible indication of some (starting)
trouble in communication and by that, as a sort of `pre-emotional',
negative state \cite{Batliner05-TOT,Steidl09-ACO}. This is corroborated by one- or two-dimensional Nonmetrical Multidimensional Scaling (NMDS) solutions, cf.\ \cite{Batliner08-PEV}, where  \textit{emphatic} is located  between \textit{neutral} and the negative states on the valence dimension.
Note that all these states, especially\ \textit{emphatic}, have only been annotated
when they differed from the (initial) neutral baseline of the speaker.

\subsection{Speech laugh and laughter}
\label{laughter}
 
 \textsc{laughter} has been annotated, together with other non-/paralinguistic events such as (filled) pauses, breathing, or (technical) noise, in the orthographic transliteration (several passes, cross-checked  by one supervisor). 
 \textsc{speech-laugh} has been annotated, together with other (prosodic) peculiarities such as unusual syllable lengthening, or hyper-correct articulation, in a separate annotation pass by one experienced labeller; 
 details are given in \cite{Steidl09-ACO}. It turned out, however, that is was necessary to re-do and correct the annotation of  \textsc{laughter} and  \textsc{speech-laugh} for the whole database; this was done by the first author.\footnote{Due to this re-labelling, the frequencies reported in this paper and in \cite{Steidl09-ACO} differ. Note that in the standard orthographic transliteration, only turns containing at least one word had been taken into account. By that, all isolated instances of \textsc{laughter} had been disregarded which are now included.} We decided not to annotate speech smile; we could only find a few somehow pronounced instances.

 The following types of laughter are annotated:   

\begin{itemize}
	\item \textsc{speech-laugh}, weak \textit{(SLw)}: speech-laugh which is not very pronounced
	\item \textsc{speech-laugh}, strong \textit{(SLs)}:  speech-laugh, pronounced
	\item \textsc{laughter}, unvoiced \textit{(Lu)}:  laughter, unvoiced throughout
	\item \textsc{laughter}, voiced-unvoiced \textit{(Lvu)}: laughter with both marked voiced and unvoiced sections; order and number of voiced and unvoiced sections are not defined
	\item \textsc{laughter}, voiced \textit{(Lv)}:  laughter, voiced throughout
\end{itemize}

%
%
 
%
%
 
 The acoustic characteristics of our laughter instances can be described along the terminology of, e.g.\, \cite{Bacharowski01-TAF,Trouvain03-SPU}: \textit{bouts}, \ie entire laugh episodes, consisting of one or several calls (segments, syllables), \ie events that clearly can be delimited. The default segmental structure of \textsc{laughter} is, as expected, [h@h@] (SAMPA notation); \textsc{speech-laugh} is often characterized by some tremolo 
which is structurally equivalent to the repetitive events in \textsc{laughter}. 
 Apart from other, more `normal' types, in a few cases, `exotic' forms such as ingressive phonation could be observed.
        
237 turns contain 276 instances of laughter, 100 \textsc{speech-laugh} and 176  \textsc{laughter}; thus each of these turns contains on average 1.16 laughter instances.
Adjacent instances of laughter and \textsc{speech-laugh} count as two separate instances.
As there are some 13.6\,k turns and some 48.4\,k words, turns with laughter instances amount to 1.7\,\% of all turns; 0.6\,\% of all tokens that are either words or laughter instances is either  \textsc{speech-laugh} or \textsc{laughter}. The approximate overall duration of the speech events in the database amounts to some 8.9 hours, the overall duration of all laughter instances to some 145 sec., \ie 0.\,4\%. 
%

To compare these frequencies with some reported in the literature: \cite{Petridis08-ALD} use seven sessions from the AMI Meeting corpus, where subjects were recruited for the task, and pre-select those 40  laughter segments that do not co-occur with speech and are ``clearly audible'' (total duration 58.4 seconds).
\cite{Kennedy04-LDI}   report ``1926 ground truth laughter events'' found in 29 meetings (about 25 hours), the so-called Bmr subset of the ICSI Meeting Recorder Corpus, divided into 26 train and 3 test meetings. \cite{Laskowski08-DOL} report for the same partition 14.94\,\% ``proportion of vocalization time spent in laughter'' for train, and 10.91\,\% for test; another subset of the ICSI meeting data (the so-called Bro subset) contains only 5.94\,\%  of laughter. This is due to different types of interaction and participants, which were more or less familiar with each other. 
On the other hand, only few laughter instances were found in ``transcript data of jury deliberations from both the guilt-or-innocence and penalty phases of 
[... a] trial'' \cite{Keyton10-ELF}:
``51 laughter sequences across 414 transcript pages''.

All these differences clearly demonstrate a strong dependency on  the scenario: on the one hand, a high percentage of laughter in scenarios where people, knowing each other quite well,  `play' meetings, having some fun, and on the other hand, children in a somehow `formal' setting, not knowing the supervisor, and trying to fulfil some tasks, or members of a jury discussing a death penalty decision.


\subsection{Syntactic boundaries}
\label{Syntactic-boundaries}

 In our scenario,  there is no real dialogue between the two partners; only the child is speaking, and the Aibo is only acting. 
 (Note, however, that there is a sort of second, marginal dialogue partner, namely the supervisor  who was present throughout the whole interaction with the Aibo. The supervisor was sometimes addressed in a sort of meta-speech, especially between the different sub-tasks, cf.\ below.)
The speaking style is rather special: there are not many `well-formed' utterances but a mixture of some long and many short sentences/chunks and one- or two-word utterances, which are often commands. Note that we have to rely on syntactic knowledge for segmenting longer stretches of speech into meaningful units. Our segmentation into turns based on a speech pause detection algorithm works sufficiently well for ASR processing; however, this procedure can result in rather long turns, up to  more than 50 words; these units are therefore not suitable for any fine-grained syntactic analysis. For the experiments presented in this study, the following  syntactic positions have been labelled by the first author; this is a sub-set of the inventory  described  fully in  \cite{Steidl09-ACO}, enriched with further `laughter-specific' positions. The inventory is based on an elaborate, shallow account of syntactic units in \cite{Batliner98-MSP}. Examples are given in parentheses for \textsc{speech-laugh} instances (typewriter font, whereas the rest of the utterances -- if any -- is given in italic). Analogously, positions of \textsc{laughter} are exemplified; English translations in italics without indication of laughter position:

\begin{itemize}
	\item   \textbf{isolated}: the turn consists only of  one instance of \textsc{laughter} 	
	\item   \textbf{vocative}:  \textsc{speech-laugh} on the vocative \textit{``Aibo''}	
	\item   \textbf{begin of unit}:  most of the time, at the begin of the turn, 
	         but can be at the begin of a syntactic unit (free phrase, clause) within a turn as well 
	         (examples: \texttt{geh} \textit{nach rechts} -- \textit{go to the right}; 
	                    \textsc{laughter} \textit{Aibo geh mal nach links} -- \textit{Aibo go to the left})
	\item 	\textbf{end of phrase}:  at the end of a free phrase, \ie a stand-alone syntactic unit but not well-formed syntactically, \ie without a verb 
	         (examples: \textit{in die andere} \texttt{Richtung} --  
	                    \textit{into the other direction};
	                    \textit{und jetzt \textsc{laughter}}-- \textit{and now})
	\item 	\textbf{end of clause}:  at the end of a main clause or a sub-ordinate clause which is syntactically well-formed 
	         (examples: \textit{was soll man da jetzt} \texttt{machen} -- \textit{what can I do now};
	                   \textit{so jetzt gibst a Ruh} \textsc{laughter} --  \textit{now keep quiet})
	\item 	\textbf{left-adjacent}:  at second position in a unit; in the first position, additionally either  \textsc{laughter}  or \textsc{speech-laugh} are found 
	         (examples: \texttt{muss ich} \textit{ihn da durch die Strassen lenken} 
	                    -- \textit{do I have to guide it through the streets};
	                    no instance of \textsc{laughter})
	\item 	\textbf{right-adjacent}:   at pen-ultima (second last) position in a unit; in the ultima (last) position, additionally either  \textsc{laughter}  or \textsc{speech-laugh} are found 
	         (examples: \textit{du musst nach} \texttt{links abbiegen} 
	                    -- \textit{you have to turn right};
	                      no instance of \textsc{laughter})
	\item 	\textbf{covering}:  the whole unit consists of \textsc{speech-laugh} with or without \textsc{laughter} 
	         (examples: \texttt{ein bisschen nach vorne} -- \textit{a little bit forwards};
	                    \textsc{laughter} \texttt{komm her} \textsc{laughter} -- \textit{come here} )
	\item  \textbf{internal}:  adjacent to the left and to the right of this label  within a unit, only words without  \textsc{speech-laugh} and no  \textsc{laughter} are found 
	         (examples: \textit{lauf} \texttt{nach} \textit{rechts} 
	                   -- \textit{go to the right};
	                    no instance of \textsc{laughter})
\end{itemize}

Note that for some syntactic positions, alternative laughter positions could be annotated: we labelled one word clauses such as \textit{``aufstehen!''} (Engl. \textit{``get up!''} with \textit{end of clause} and not with \textit{covering}.  With \textit{covering}, all words belonging to a unit were annotated even if of course, one of them is at the beginning and one in ultima position. These decisions are somehow arbitrary; in no case, however, the interpretation in Sec. \ref{Syntax} would change if the one or the other decision had been taken.


\section{Results}
\label{Results}

Our data are not Gaussian-distributed, and several outliers such as speaker-specific frequencies or extreme duration values can be observed. Thus we decided in favour of non-parametric statistic procedures such as Spearman's rank co-efficient, Chi-Square, and Mann-Whitney U-test; p-values reported are always for the two-tailed test. 
Note that `significant' p-values should rather be taken as indicating `large enough differences' in the sense of \cite{Eysenck60-TCO,Rozeboom60-TFO}, and not in the strict sense of inferential statistics. 
We therefore  refrain  from using adjusted levels of significance. (For our results, using them would simply mean only to treat p-values below 0.01 as `significant'.)

\subsection{Speaker- and gender-specific use of laughter}
\label{Speaker-and-gender}

\begin{figure} [!tb]
  \centering
  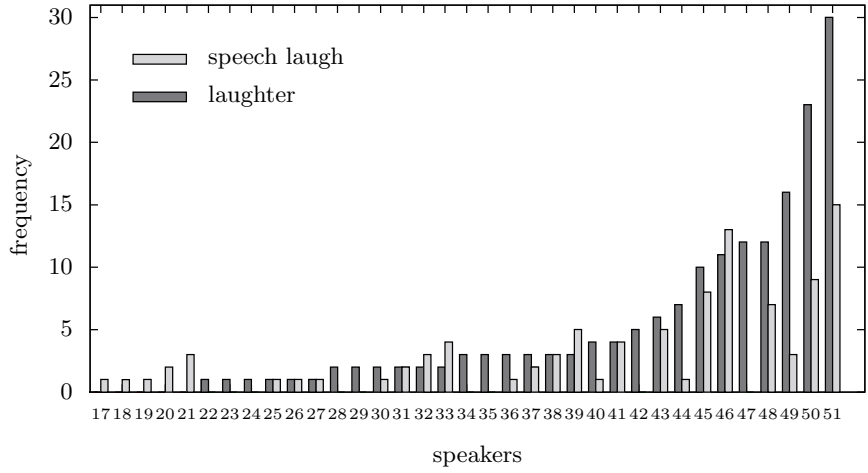
  \caption{Distribution of \textsc{laughter}  and  \textsc{speech-laugh} amongst speakers,
           sorted by frequency of  \textsc{laughter} per speaker
}
  \label{distribution-speakers}
\end{figure}

 Figure \ref{distribution-speakers}  displays the speaker-specific frequencies of   \textsc{laughter} and  \textsc{speech-laugh},  sorted by frequency of  \textsc{laughter} per speaker; 16 speakers, \ie almost one third,  are omitted because they produced neither   \textsc{laughter} nor \textsc{speech-laugh}.
 The  Spearman rank co-efficient between \textsc{laughter} and  \textsc{speech-laugh} is 0.61 if all speakers, even those that did not produce  \textsc{laughter} or \textsc{speech-laugh}, are taken into account, and 0.43 if only speakers that produced at least one instance of \textsc{laughter} or \textsc{speech-laugh} are processed. Thus, some tendency can be observed to display either no laughter or both types of laughter. 
We can assume that the decision between `to laugh or not to laugh' is grounded in some basic attitude towards the task and towards the situation as a whole, which in turn might be caused by differences in the character of the children \cite{Batliner08-PEV}. In a Mann-Whitney test, there is no significant gender difference as for absolute frequencies of \textsc{laughter} and \textsc{speech-laugh}, their different sub-types specified below, or their frequencies relative to absolute word frequencies per speaker. With other words: at least in this setting, girls and boys seem not to differ in their use of laughter, cf.\ below Section \ref{Syntax} as well.

\subsection{Duration of laughter}
\label{Duration}

%
%

\begin{table}[htb]
  \renewcommand{\baselinestretch}{1}          
\small\normalsize  
\caption{Duration of types of laughter in \# frames (10 msec.)}
\label{Duration-of-laughter}
\vspace{0.3cm}
\centering
\begin{tabular}{|l|r|r||r||r|r|r||r|} \hline
 statistics      &  \textit{SLs}  &  \textit{SLw}  &  \textit{SLtot}&  \textit{Lv}  &  \textit{Lvu}  &  \textit{Lu}  &  \textit{Ltot} \\ \hline
\# tokens        &  44      &  54      &  98     &  32      &  69      &  75      &  176     \\ \hline\hline
Mean             &  54.84 &  45.10 &  49.49 &  53.00 &  63.79 &  46.13 &  54.30 \\
Median           &  55 &  44 &  49.50 &  44 &  47 &  39 &  42.50 \\
Std. Deviation   &  20.14&  22.37&  21.84&  37.02&  58.61&  49.74&  51.85  \\
Skewness         &  .31    &  .57    &  .37    &  1.93   &  3.32   &  5.78   &  4.16   \\
Minimum          &  21   &  6    &  6    &  19   &  5    &  9    &  5    \\
Maximum          &  100  &  113  &  113  &  173  &  328  &  414  &  414  \\ \hline
\end{tabular}
\vspace{0.5cm}
\end{table}

Table \ref{Duration-of-laughter} displays some statistical key figures for the five types of laughter separately, and for the two main types \textsc{speech-laugh} \textit{(SLtot)} and \textsc{laughter} \textit{(Ltot)}. All these distributions are skewed right (row `skewness') which could be expected, as it is duration data. \textit{SL}-types, \ie words, are less skewed than \textit{L}-types, and amongst these latter ones, those containing unvoiced parts are skewed most. This is due to a few outliers, \ie very long \textsc{laughter} instances: the median is more uniform across the types than the mean (and by that, standard deviation and maximum values). In a Mann-Whitney test,  the durations of  \textit{SLs} vs.\ \textit{SLw} differ with ($p=0.024$). This can of course be due to differences in word length but most likely, to \textit{SLs} being more pronounced and by that, longer, than \textit{SLw}. Three pair-wise Mann-Whitney tests resulted in one of the differences, namely \textit{Lvu} vs.\ \textit{Lu}, being significant with ($p=0.001$). This might be due to two factors: \textit{Lu} tends to be weaker and by that, shorter, and for \textit{Lvu}, the alternation of voiced and unvoiced might automatically `result' in some longer duration.

\subsection{Syntactic position of laughter}
\label{Syntax}

\begin{figure} [!tb]
  \centering
  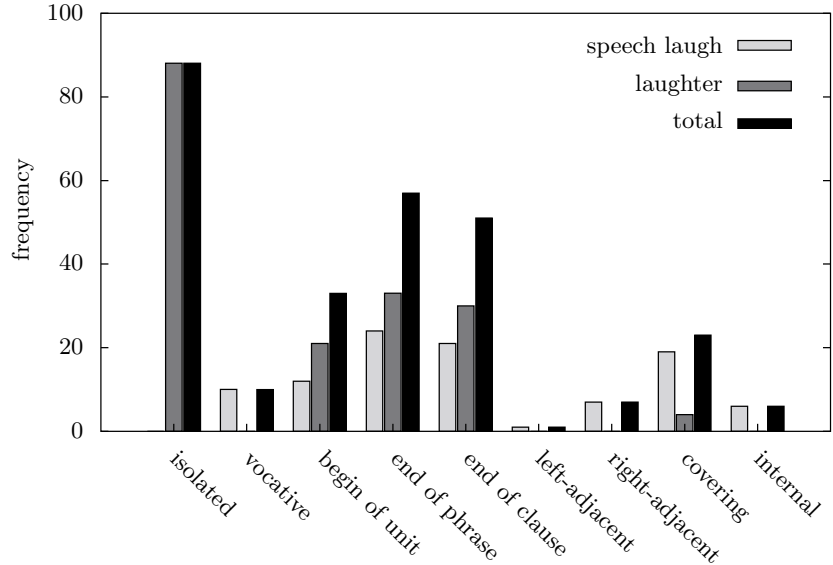
  \caption{Distribution of types of laughter for syntactic positions
}
  \label{SL-L}
\end{figure}

Figure \ref{SL-L} displays the frequencies of \textsc{speech-laugh} and \textsc{laughter} for all different syntactic positions. These are absolute figures; note that there are 21 male and 30 female children in our database. A Mann-Whitney test resulted in no significant differences  ($\alpha=0.5\,\%$)  between the genders, separately for each syntactic position or for the totals of \textsc{speech-laugh}, \textsc{laughter}, or both taken together. 
Thus it seems save to conclude that both girls and boys employ laughter the same way.

%

%
%
%


In Figure \ref{SL-L},  it is trivial that in \textit{isolated} position, only \textsc{laughter} occurs because one-word turns constitute by definition either a clause or a free phrase. The same way, it is trivial that \textit{vocatives} only constitute \textsc{speech-laugh}. There is not much difference between \textsc{speech-laugh} and \textsc{laughter} as for \textit{begin of unit},  \textit{end of phrase}, and   \textit{end of clause}. However, if we map these instances onto a main class \textit{edge position}, then the difference between 57  instances of \textsc{speech-laugh} and 84 instances of  \textsc{laughter} is significant in a chi-square test ($p=0.023$). For the complement,
 \ie  \textit{adjacent/covering/internal} positions,
the difference between 33 instances of  \textsc{speech-laugh}  and 4 instances of \textsc{laughter}  -- which are all in ultima position, cf.\ below -- is more marked, with ($p=0.000$).

 \textit{left/right-adjacent}  means an internal second or pen-ultima position but only when the first or ultima position is \textsc{laughter} or  \textsc{speech-laugh}: there is none for \textsc{laughter} and only one and seven, respectively, for \textsc{speech-laugh}.  \textit{covering} means that throughout a syntactic constituent, there is \textsc{laughter} and/or \textsc{speech-laugh}; note that all four instances of \textsc{laughter} in  \textit{covering} occur in first or ultima position. Thus there is no \textsc{laughter} in   \textit{internal} position, only \textsc{speech-laugh}.  These distributions mean that \textsc{laughter} can only be observed  \textit{isolated}, \ie marked-off from speech, or at the beginning/end of syntactic constituents, and that \textsc{speech-laugh} is either -- and most of the time -- at the fringe (mostly at the end) of syntactic constituents, or coherent left/right-adjacent or covering but very seldom  stand-alone internally in a syntactic constituent. 

In \cite{Provine93-LPS} it is claimed that \textsc{laughter} sort of punctuates speech, \ie it is almost always found at those positions where we punctuate in written language.
 This turns out to hold true for our data as well: \textsc{laughter} is never internal. In contrast, we see that \textsc{speech-laugh} can occur in internal position, but very seldom. 
In the overwhelming majority of the cases, \textsc{speech-laugh} is found at the edges of linguistic units -- but these can be found within turns, \ie longer stretches of speech, as well. 

To our knowledge, there are not many studies on the relationship between human speech/linguistic processing and paralinguistic processing -- such as laughter.  We know, however, that phonetic/psycholinguistic studies on the localisation of non-verbal signals within speech showed that listeners tend to structure the perception of these phenomena along the perception and comprehension of linguistic phenomena (sentence processing), cf.\ \cite{Garrett66-TAU}.\footnote{In \cite{Provine93-LPS}, only \textsc{laughter} and not  \textsc{speech-laugh} is addressed. A weak point of this study might be that the data were annotated online by `observers' of anonymous subjects in public places. Thus the localisation of \textsc{laughter} could not be checked later on. It might be that these observers displayed the same tendency to localise such events at syntactic boundaries, even if they are not. However, our findings point towards the same direction.
}  
Our findings suggest a somehow close relationship between both phenomena -- otherwise, there would be no reason why we should not observe laughter in internal position.  Thus, it might be that linguistics and paralinguistics are not just two independent streams but  more intertwined, cf.\ \cite{Nwokah99-TIO}, p. 892: ``A view of laughter as merely a suprasegmental overlay on the segmental organization of speech is clearly an inadequate view of speech-laugh patterns. Unlike stress or intonation, laughter can stand independently as a meaningful communicative response.'' But, we have to add, it is embedded in syntactic structure as well.

\subsection{Communicative function of laughter}
\label{Communication}

276 laughter instances (\textsc{speech-laugh} and \textsc{laughter}) are found in 237 turns, \ie 1.16 on average per turn. 88 are isolated \textsc{laughter} instances, constituting a turn; 40 are `meta-statements', not directed towards the Aibo but being either private speech (directed to one-self) or directed towards the supervisor; these meta-statements can be conceived as constituting `off-talk' \cite{Batliner08-TTO}. Sometimes, it is not easy to tell these different types apart: the exclamation  \textit{``s\"{u}{\ss}!''} (Engl. \textit{``sweet!''}) could be both, \textit{``passt das so?''} (Engl. \textit{``is that ok?''}) is clearly directed towards the supervisor. Thus, some 46\,\% of the turns containing laughter instances constitute interactions (mostly the illocution `command') directed towards the Aibo, some 54\,\% of these turns do not belong to any interaction with the Aibo. In both constellations, laughter can -- but need not be -- an indication of emotion (emotional user-state); obvious would be the indication of  \textit{joyful}.

%
%

\begin{table}[htb]
  \renewcommand{\baselinestretch}{1}          
\small\normalsize  
\caption{Cross-tabulation of word based emotion labels with \textsc{speech-laugh}}
\label{emotion-speech-laughter}
\vspace{0.3cm}
\centering
\begin{tabular}{|l|r|r|r|} \hline
emotion label &  \textit{SLs}     & \textit{SLw}  &  sum  \\ \hline
 \textit{mixed}         &   8    &   8    &  16    \\
 \textit{angry}         &   1    &   1    &   2    \\
 \textit{joyful}        &   25   &  24    &  49    \\
 \textit{neutral}       &   10   &  21    &  31     \\ \hline
total         &  44    &  54    &  98   \\ \hline
\end{tabular}
\vspace{0.5cm}
\end{table}

Table \ref{emotion-speech-laughter} displays the word-based co-occurrence of \textsc{speech-laugh} and emotion label. Indeed, in 50\,\% of the cases, \textsc{speech-laugh} obviously contributes to the indication of  \textit{joyful}; these 49 cases amount to 49\,\% of all 101 \textit{joyful} instances. 32\,\% of the words are  \textit{neutral}, 16\,\% belong to a \textit{mixed} rest class,\footnote{\textit{mixed} means that in these cases, no majority label could be given; this is different from the \textit{rest} class in Section \ref{Emotion}.}  2\,\% to  \textit{angry}, and no case to  \textit{motherese}. At first sight, this could not be expected, when we consider the literature: \cite{Nwokah99-TIO} showed that in mother-child interaction, mothers produced some 50\,\%  \textsc{speech-laugh}. This \textit{child-directed} speech has different names such as  \textit{motherese, parentese}, or \textit{register of intimacy,} with somehow different connotations but having a great deal in common with our  \textit{motherese} cases, cf.\ \cite{Batliner06-TPO,Batliner08-MAC}, such as lower harmonics-to-noise ratio, lower energy, and at the same time, more variation in energy and $F_0$. Obviously, laughter is no common trait:
in a mother-child (\ie mother-baby) interaction, the
eliciting of social (mutual)  laughter serves as reinforcement  of a good
parent-child attachment, and as confirmation of the child's well-being.
In our scenario, laughter is not social in this sense, and is not used to
establish specific relationships with the Aibo. In fact, any attempt to elicit laughter
would be in vain because the Aibo is simply not programmed that way.

Similar results are obtained when we compare the speaker-specific frequencies of  \textit{motherese},  \textit{angry}, all words, \textsc{speech-laugh} and \textsc{laughter} in Table \ref{correlation-speaker}: there are significant -- albeit not very high --  positive correlations between  \textit{motherese},  \textit{angry}, and all words. This means that subjects show some variability but do not have any bias towards a positive or negative attitude. The same way, there is a significant correlation between the frequencies of \textsc{speech-laugh} and \textsc{laughter}: subjects seem not to prefer strongly the one or the other type of laughter. 
There are, however, very low correlations between any type of laughter and any type of emotion -- apart from  \textit{joyful}. 
This might show that this child-robot relationship is peculiar, and really half-way between `close and intimate' and `distant and not intimate'. This can be traced back to the distance (the child is some 1.5\,m away from the robot), to the robot being a robot, to the robot being a `tin-box' and not a sweet furry baby seal, etc. -- we do not know yet. Anyway, the assumption of \cite{Nwokah99-TIO}, p. 892 that ``... speech-laughs are only common in particular social contexts, such as maternal infant-directed speech during play or in the laughter of close relationships'' has to be specified: speech-laughs can be observed in other settings; however, they might be not as frequent. The `normal', default function of the laughter instances found in our database really seems to be an indication of amusement or joy. Some other functions are detailed in the next section.


\begin{table}[htb]
  \renewcommand{\baselinestretch}{1}          
\small\normalsize  
\caption{Correlations (Spearman) for speaker-specific frequencies; 51 speakers; correlations with `*' are significant at $\alpha = 0.001$ }
\label{correlation-speaker}
\vspace{0.3cm}
\centering
\begin{tabular}{|l|c|c|c|c|c|} \hline 
type           &    \textit{~~motherese~~}  &   \textit{~~~~~angry~~~~~}  
&  ~~~~~words~~~~~  &  \textsc{speech-laugh}   
\\  \hline
 \textit{motherese}      &       --       &         &         &              \\
 \textit{angry}          &      .53*     &    --    &         &              \\
words          &      .51*     &   .53*  &   --     &              \\
\textsc{speech-laugh} &  .10   &   -.23  &  -.05   &   --           \\
\textsc{laughter}     &      .14      &   -.15  &  .20    &   .60*       \\ \hline
\end{tabular}
\vspace{0.5cm}
\end{table}

\subsection{Temporal position of laughter in the dialogue}
\label{Position}

\begin{figure} [!tb]
  \centering
  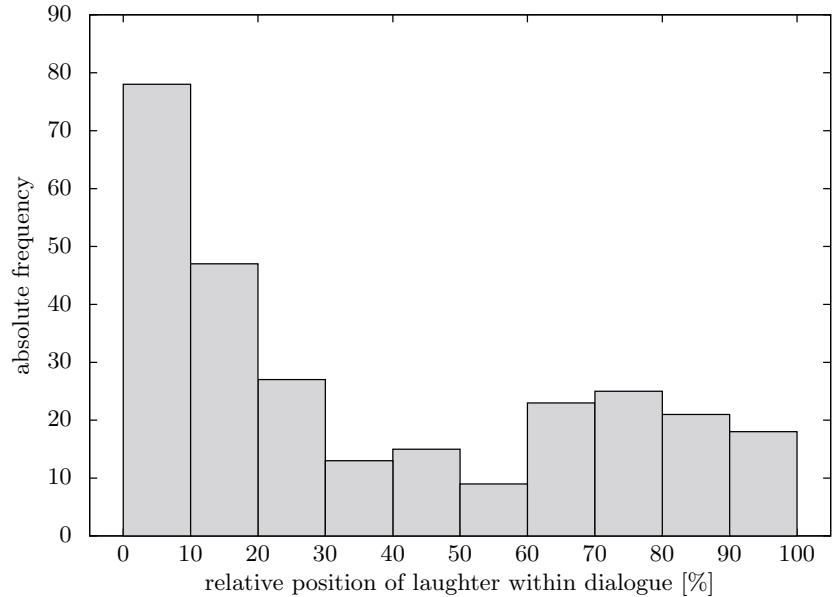
  \caption{Frequencies (y-axis) of relative position of laughter in the time course of the dialogue, 
               smoothed by quantisation into 10 percentile slides (x-axis)
}
  \label{position}
\end{figure}

 In Sec. \ref{Syntax} we have seen that laughter is very often  found at a  syntactic edge position, \ie either at the begin or at the end of syntactic units. We now want to have a look at the position of laughter in the whole dialogue: the children had to complete five tasks, three short ones, one longer, and again, two short ones, cf.\ Sec. \ref{Database}. It is likely that a child exhibits the same linguistic behaviour throughout the whole communication: they are either talkative or not.  Remember that  Aibo's actions were predefined and did not depend on the children's commands.    We therefore computed for each laughter instance its relative position in the dialogue by dividing the turn number the laughter belongs to with the maximal number (number of the last turn in the dialogue.) To get a somehow clear picture without outliers, we truncated the resulting figures, aiming at a quantisation into 10 percentile slides, cf.\  Figure  \ref{position}. Thus the positions given in the figure denote approximately the temporal position in whole interaction, \ie in the task structure. We can see a first maximum at the beginning of the dialogue, then a descending  slope and later on, a second weak maximum between 60\,\% and 80\,\% of the dialogue. Note that in a few cases, there are some `artifacts' because not the whole communication was recorded due to technical problems. Moreover, some children do not produce any laughter, and a few other ones quite a lot. Thus we have to interpret this outcome with due care: 
the first maximum at the beginning  of the whole interaction can be interpreted as `phatic' laughter -- most of the time, isolated  \textsc{laughter}, most likely denoting something like joyful expectation, mixed with tension. 
Then, the descending slope might indicate some focusing on the task.  Later on, when the children are more familiar with the tasks and the supervisor, 
laughter can often be found  at the transition of one task to the following  task; here, some children made meta-statements, including laughter, such as \textit{`` ... sometimes, it doesn't want to listen at all \textsc{laughter}''}. We might suppose that laughter is influenced  by antagonistic tension and relief; it occurs more often at the beginning as well as at the end of tasks than in the `normal' course of the tasks. A `punctuation' function can thus be observed at the lower level of syntactic structure as well as -- in a less restricted way -- at the higher level of dialogue structure. 
 
%
%
%
%

%

\section{Automatic classification of laughter}
\label{Automatic-classification}


For the automatic classification of laughter, state-of-the art audio signal processing and classification methods are used. 
A large set of acoustic features is extracted from segments (complete turns, or single word units) of the audio recording. In total the feature set  considered contains 5\,967 acoustic features, which are extracted using the openEAR framework~\cite{Eyben09-OIT}. Thereby, for each input segment of variable length, a static feature vector is extracted by applying various functionals to low-level feature contours. The latter include low-level descriptors (LLD) such as signal energy, Mel-frequency cepstral coefficients (MFCC), fundamental frequency, probability of voicing, zero-crossing rate (ZCR), etc., and their respective first and second order regression coefficients.
The 39 low-level descriptors are summarised in Tab.\ \ref{tab:lld}. A list of the 51 functionals can be found in Tab.\ \ref{tab:functionals}. 
These and similar features have been applied successfully to various audio classification tasks, e.g.\ musical genre recognition~\cite{Schuller06-MST}, emotion recognition~\cite{Grimm07-AER,Schuller07-TRO}, and classification of non-linguistic vocalisations~\cite{Schuller08-SAD}.

\begin{table}[htb]
\centering
\begin{tabular}{l|p{0.80\columnwidth}}
\hline\hline
\textbf{LLD Group}		& \textbf{Description} \\
\hline
Time Domain			& Zero-crossing rate (ZCR), max./min.\ sample value, signal offset\\     				
Energy 				& Root mean-square (RMS) \& logarithmic \\ 								
Voice					& Fundamental frequency $F_0$	via autocorrelation function (ACF)\\
						& Probability of voicing ($\frac{\mathrm{ACF}(T_0)}{\mathrm{ACF}(0)}$)	\\ 	
						& $F_0$ quality ($\frac{\mathrm{ZCR(ACF)}}{F_0}$)	\\ 	
		  			 	& Harmonics-to-noise ratio (HNR) \\									
Spectral				& Energy in bands 0\,-\,250\,Hz, 0\,-\,650\,Hz, 250\,-\,650\,Hz, 1\,-\,4\,kHz \\	
						& 10\,\%, 25\,\%, 50\,\%, 75\,\%, and 90\,\% roll-off points,
						 centroid, flux, and relative position of spectral maximum and minimum \\ 
Cepstral				& MFCC 0-15 \\															
\hline\hline
\end{tabular}
\vspace{2mm}
\caption{39 acoustic low-level descriptors (LLD) for generation of a large acoustic feature set.}
\label{tab:lld}
\end{table}

\begin{table}[htb]
\centering
\begin{tabular}{p{0.85\columnwidth}|c}
\hline\hline
\textbf{Type} & \textbf{\#} \\
\hline
Max.\,/\,min. value and respective relative position & 4 \\
Range (max.\,-\,min.), max.\,/\,min. value - arithmetic mean & 3 \\
Arithmetic and quadratic mean  & 2 \\
Arithmetic mean of absolute and non-zero values & 2 \\
Percentage of non-zero values in contour & 1\\
Quartiles, inter-quartile ranges, 95\,\% and 98\,\% percentiles & 8 \\
Standard deviation, variance, kurtosis, skewness & 4 \\
Centroid of feature contour & 1 \\
Zero-crossing and mean-crossing rate & 2 \\
25\,\% down-level time, 75\% up-level time & 2 \\
Rise-time, fall-time & 2\\
Number of peaks, mean distance between peaks & 2\\
Mean of peaks, mean of peaks - arithmetic  mean & 2\\
Number of segments based on $\Delta$-thresholding & 1 \\
\hline
Linear regression coeff.\ and corresponding approximation error & 4 \\
Quadratic regression coeff.\ and corresponding approximation error & 5 \\
\hline
Discrete cosine transformation (DCT) coefficients 0-5 & 6 \\
\hline
\hline
\end{tabular}
\vspace{2mm}
\caption{51 functionals (statistical, polynomial regression, and transformations) applied to low-level descriptor contours.}
\label{tab:functionals}
\end{table}

In order to find a set of features highly relevant for laughter classification, an automatic data-driven feature selection method called \textit{correlation-based feature-subset selection} (CFS)~\cite{Witten05-DMP} is used. This method evaluates the relative importance of features based on their correlation to the class. The method is independent of the type of classifier that will be used to do the actual recognition work. This is both an advantage and disadvantage of CFS. Due to not including the classification method into the feature selection procedure, the obtained feature sub-set is very likely to be suboptimal for the chosen classification method. However, this approach leads to a more general feature set when 
we try to interpret the selected features as characteristic acoustic traits of laughter.
Two different levels of input segmentation are investigated, which correspond to  different practical applications: \textit{turn-based} laughter detection, and \textit{word-based} laughter classification.\footnote{This is a somehow sloppy word usage of `detection' in order to tell apart these two different tasks: we `detect' that laughter occurs somewhere in the turn but we do not localise it; in `classification', localisation is given and we decide whether an item belongs to the one or the other class(es).} 

For turn-based laughter detection, a single acoustic feature vector is extracted from the full length input turn. A two-class decision is performed, \ie whether the turn contains laughing (of any type) or not. The classes \textit{L} and \textit{W} are assigned to the turns respectively. This scenario is of a highly practical relevance, since in most cases, systems need to know whether a person is laughing or not, while the exact position of the laughter within the utterance is irrelevant. Furthermore, a three-class decision is analysed where we discriminate between turns containing no laughing at all (\textit{W}), \textsc{speech-laugh} only (\textit{SL}), and \textsc{laughter} (\textit{L}) (possibly mixed with \textsc{speech-laugh}).

For word-based laughter classification, the turns were manually segmented into word units. A word unit thereby spans exactly one word or a non-linguistic vocalisation, such as isolated laughter. Features are extracted per word unit segment.
 Each word unit is assigned one of the six classes, word (\textit{W}), weak \textsc{speech-laugh} (\textit{SLw}), strong \textsc{speech-laugh} (\textit{SLs}), voiced \textsc{laughter} (\textit{Lv}), mixed voiced and unvoiced \textsc{laughter} (\textit{Lvu}), and unvoiced \textsc{laughter} (\textit{Lu}). Since some of these six classes might not be clearly distinguishable, they have been combined for additional experiments. This results in two sets of labellings, one containing two labels (word \textit{W}, and \textsc{laughter} \textit{L}), and the other containing three labels (word \textit{W}, \textsc{speech-laugh} \textit{SL}, and \textsc{laughter} \textit{L}). Even though word-unit based laughter classification requires a segmentation into word-units, which cannot be done perfectly automatically, it is interesting from a research point of view. A comprehensive study of relevant acoustic properties of speech vs.\ laughter can be conducted by automatically analysing acoustic features relevant for automatic classification.





In order to obtain speaker independent classification results on the whole FAU Aibo corpus, \textit{leave-one-speaker-out cross-validation} is performed. Thereby, evaluation is performed in 51 folds, corresponding to the 51 speakers in the corpus. In each fold the data from one speaker is used as a test set, while the data from the remaining 50 speakers is used for training and feature selection. 

Since significantly more word-units and turns without laughter are found in the FAU Aibo corpus, the data set is highly unbalanced between the speech class on the one hand and the \textsc{laughter} and  \textsc{speech-laugh} classes on the other hand. Therefore, for training robust classification models which do not show a bias towards the speech class, it is necessary to create a balanced training set for the training phase. 
We therefore limit the number of training instances in the speech class for word-unit based experiments 
to 100 (6-class problem) and to 300 (for the 2-class and the 3-class problem), and for turn based experiments to 237 (2-class problem) and to 150 (3-class problem) by random sub-sampling of speech class instances. Note that for the turn-based three class problem, the number of  \textsc{speech-laugh} instances was also limited to 150 by random sub-sampling.


Feature selection using CFS is performed independently for each fold on the respective training set only.  Thus, we ensure that the respective test set is completely unknown to the system with respect to both feature selection and model; this would not be the case if we performed the feature selection on the whole corpus. 
However, the proposed method results in 51 \textit{different} feature sets, each containing approximately 100--200 selected features, specific to the respective training partition. In order to find overall relevant features, we create a new feature set by including only those features which have been selected in all 51 folds. This method yields even smaller sets of features (approx.\ 20--30 features). Better classification is obtained with these feature sets than with the individual per-fold feature sets. In some cases the performance is even superior to using the full feature set. This indicates the high relevance of these selected features, as will be discussed later. 

As classifier we use Support-Vector Machines (SVMs) as described in~\cite{Witten05-DMP}. SVMs have shown excellent performance for related tasks, e.g.\ classification of non-linguistic vocalisations (e.\,g.~\cite{Schuller08-SAD}), and emotion recognition (e.\,g.~\cite{Vlasenko07-FVT}).


\subsection{Classification performance}


A summary of all results obtained for automatic laughter detection and classification is shown in Tab.\ \ref{results-overview}. We see that generally, turn-based classification outperforms word-based classification. This might be at least partly due to the fact that there is on average 1.16 instances of laughter and by that, more than one `island of markedness'  per turn.
Of course, the more detailed 3-class and the 6-class problem result in lower classification performance.
Both for weighted average recall (WA)  and  unweighted average recall (UA)\footnote{
WA is the overall recognition rate or recall (number of
correctly classified cases divided by total number of cases); UA is the `class-wise' computed recognition rate, \ie the mean along the diagonal of the confusion matrix in percent.}, 
classification performance is better if using all features ($FS_{n}$) than if using features selected in all 51 folds ($FS_{c}$), and both procedures are better than if using features selected via CFS on the respective full set, \ie data from all 51 folds combined ($FS_{f}$); cf.\ our remarks above  on the advantages and disadvantages of CFS. There are two exceptions if we look at UA for $FS_{c}$  vs.\  $FS_{f}$  for  the 3-class and the 6-class word based problem: obviously, the more detailed the task, the more features have to be employed for modelling and classifying the classes.

\begin{table}[th]
  \caption{Overview of all results for turn and word based segmentation with different number of classes: $N_{cl} = $ 2, 3, and 6 classes. 51-fold leave-one-speaker-out cross-validation. \textit{Dummy}: Correctly classified instances by always choosing the most likely class as seen in the training set distribution. Classification using all features ($FS_{n}$), 
features that have been selected in all 51 folds ($FS_{c}$), and features selected via CFS on the respective full set, \ie data from all 51 folds combined ($FS_{f}$). 
Number of features that was selected in all 51 folds ($N_{c}^{ft}$), and number of features selected on the full FAU Aibo set ($N_{f}^{ft}$). Training on near balanced set (see text), evaluation on full set (highly unbalanced). Weighted average recall (WA) and, in parentheses, unweighted average recall (UA).}
    \label{results-overview}
\vspace{0.3cm}
  \centerline{
 \begin{tabular}{|l|cc|c|cc|cc|}  \hline
[\,\% WA (UA)]   & $N_{cl}$ & \textit{Dummy} & $FS_{n}$ & $FS_{c}$ & $N_{c}^{ft}$ & $FS_{f}$ & $N_{f}^{ft}$\\
\hline
\textbf{Turn} & & & & & & & \\
     & 2          & 50.0  (50.0)     & 84.7 (85.0)        &   82.0 (83.0)    &   30  & 80.9 (82.2) & 156  \\
     & 3          & 41.1  (33.3)     & 81.8 (89.5)        &   77.5 (64.5)    &   20  & 71.9 (60.9) &  121 \\
\hline
\textbf{Word} & & & & & & & \\
     & 2          & 52.3  (50.0)     & 77.9 (78.5)        &    76.6 (77.5)   &   25  & 74.8 (76.0) & 176 \\
     & 3          & 52.3  (33.3)     & 77.6 (69.8)        &    73.7 (64.3)   &   34  & 67.4 (71.2) & 183 \\
     & 6          & 26.7  (16.7)     & 58.3 (49.8)        &    54.6 (41.3)   &   13  & 52.2 (50.1) & 161 \\
\hline
 \end{tabular}
}
\end{table}

Tab.\ \ref{Classification6-selection} shows the confusion matrix for the 6 class word-unit based classification problem using a set of only 13 features 
constituting the intersection of selected features in all 51 folds, cf.\ Tab. \ref{results-overview}, last row.
It shows confusions that are expected from a phonetic viewpoint: confusions within the classes   \textsc{laughter}  and \textsc{speech-laugh}  are more frequent than confusions between \textsc{laughter} and \textsc{speech-laugh}. Moreover, words are often misclassified as \textsc{speech-laugh}, where weak \textsc{speech-laugh} is more frequent than strong \textsc{speech-laugh}, which is to be expected. 
There is some misclassifications of words as \textsc{laughter}, especially \textit{Lv}. The confusions with \textit{Lu} can be explained by the fact that especially shorter words can either be unvoiced or the feature extraction might not have given reliable results.

%
%
%
%

\begin{table}[th]
  \caption{Word-based classification with 51-fold leave-one-speaker-out cross-validation; with 13 features selected; 
  54.6\,\% correct; 6-class problem}
    \label{Classification6-selection}
\vspace{0.3cm}
  \centerline{
  \begin{tabular}{|l|r|rr|rrr||r|r|}  \hline
class. as  & \textit{W}   & \textit{SLw}    & \textit{SLs}  & \textit{Lu}    & \textit{Lvu}   & \textit{Lv}  & \#   & \,\% corr.\\ \hline
\textit{W}		& \textbf{412}	& 117	& 76	& 25	& 24	& 51 & 	705        &   58.4  \\ \hline
\textit{SLw}	& 23	& \textbf{5}	& 16	& 5	& 3	& 2 & 54                  &   9.3  \\ 
\textit{SLs}& 7	& 15	& \textbf{15}	& 2	& 3	& 2	&  44                   &  34.1   \\ \hline
\textit{Lv}		& 10	& 4	& 2	& \textbf{2}	& 12	& 2	& 32                &   6.3  \\ 
\textit{Lvu}& 2	& 4	& 3	& 0	& \textbf{44}	& 16 & 69                     &  63.8  	\\ 
\textit{Lu}		& 2	& 0	& 2	& 2	& 12	& \textbf{57} & 75                  &  76.0 	\\ \hline
\end{tabular}
}

\end{table}
Tab.\ \ref{Classification3-selection} shows the confusion matrix for the 3 class word-unit based classification problem using a set of 34 features determined by automatic feature selection of commonly selected features in all 51 folds.
\begin{table}[th]
  \caption{Word-based classification with 51-fold leave-one-speaker-out cross-validation;
   with 34 features selected; 73.7\,\% correct; 3-class problem} 
    \label{Classification3-selection}
\vspace{0.3cm}
  \centerline{
  \begin{tabular}{|l|r|rr||r|r|}             \hline
class. as & \textit{W}	& \textit{SL}	& \textit{L}	& \#              &  \,\% corr. \\ \hline 
\textit{W} 		& \textbf{552}	& 77	& 76	& 705   & 78.3  \\ \hline
\textit{SL}		& 44	& \textbf{40}	& 14   & 98     & 40.8  \\ 
\textit{L}		& 25	& 21	& \textbf{130}	& 176     & 73.9  \\ \hline
\end{tabular}
}
\end{table}
This table shows a good recognition performance for the classes word and \textsc{laughter}. \textsc{speech-laugh}, however, is often confused with words, which shows the challenge of detecting laughter in speech.\footnote{Note that out of the 100 instances of \textsc{speech-laugh}, 2 cases could not be processed because they were too short to extract meaningful features based on functionals.} More \textsc{speech-laugh} word-units are classified incorrectly as words than are classified correctly. Looking once more at the result in Tab.\ \ref{Classification6-selection}, where we can see that weak  \textsc{speech-laugh}    and words are likely to be confused, we can assume that those weak \textsc{speech-laugh} word-units, which are now combined with the strong   \textsc{speech-laugh} instances, lead to the poor performance for the overall \textsc{speech-laugh} class. 


Tab.\ \ref{Classification3t-selection} shows the confusion matrix for the 3-class turn based 
detection
problem using a set of 20 features determined by automatic feature selection of commonly selected features in all 51 folds.
Tab.\ \ref{Classification2t-selection} shows the confusion matrix for the 2-class turn based
 detection 
 problem using a set of 30 features determined by automatic feature selection of commonly selected features in all 51 folds. 3-class turn-based laughter 
detection 
 shows similar results to the 3-class word-based laughter classification, where performance for the \textsc{speech-laugh} class is weak. Restricting the 
 problem to a binary speech/laughter decision improves the total number of \textit{W}
 instances classified correctly as \textit{W}
 and at the same time increases the number of \textit{L}
 instances correctly recognised from 163 (when combining correctly recognised \textsc{speech-laugh}   and \textsc{laughter}) to 199.

\begin{table}[th]
  \caption{Turn-based 
                    detection with 51-fold leave-one-speaker-out cross-validation; with 20 features selected; 77\,\% correct; 3-class problem}
    \label{Classification3t-selection}
\vspace{0.3cm}
  \centerline{
  \begin{tabular}{|l|r|rr||r|r|}  \hline
class. as & \textit{W}	& \textit{SL}	& \textit{L}	  & \#                  & \,\% corr.     \\ \hline
\textit{W}	   & \textbf{10475}	& 1486	& 1533  &  13494  & 77.6   	\\ \hline
\textit{SL}	   & 23	& \textbf{22}	& 20  &  65	          & 33.8    \\ 
\textit{L}	   & 15	& 16	& \textbf{141}	&  172          & 82.0    \\ \hline
\end{tabular}
}
\end{table}
%

%
\begin{table}[th]
  \caption{Turn-based 
    detection with 51-fold leave-one-speaker-out cross-validation; with 30 features selected; 82\,\% correct; 2-class problem}
    \label{Classification2t-selection}
\vspace{0.3cm}
  \centerline{
  \begin{tabular}{|l|rr||r|r|}  \hline
class. as   & \textit{W}	& \textit{L}	&  \#                 & \,\% corr.  \\ \hline
\textit{W}			& \textbf{11054}	& 2440	& 13494   &  81.9  \\ 
\textit{L}			& 38	& \textbf{199}	& 237         &  84.0 \\ \hline
\end{tabular}
}
\end{table}


\subsection{Feature interpretation}




The common features selected in all 51 folds for at least two experiments of turn or word-unit based 
detection/classification  with 2, 3, and 6 classes for word-unit based classification (W$_2$, W$_3$, W$_6$), and 2 and 3 classes for turn based detection (T$_2$, T$_3$) are shown in Tab.\ \ref{tab:commonfeatures}. The check marks show for which of the experiments the features were selected in all 51 folds. 
It is notable that the features selected for turn based and word-unit based detection/classification are almost completely disjoint. Only two features were selected for at least one turn- and word-unit based detection/classification experiment: the zero-crossing rate of $\Delta$ MFCC$_2$ and the inter-quartile range 3-1 of the spectral flux.

\begin{table}[th]
  \caption{Common features selected in all 51 folds of at least two experiments for turn and word-unit based detection/classification. 2, 3, and 6 classes for word-unit based classification (W$_2$, W$_3$, W$_6$), and 2 and 3 classes for turn based detection (T$_2$, T$_3$).}
    \label{tab:commonfeatures}
\vspace{0.3cm}
\centering
  \begin{tabular}{p{0.7\columnwidth}|ccc|cc}  \hline\hline
\textbf{Feature description}            & W$_2$ & W$_3$ & W$_6$    & T$_2$ & T$_3$ \\ \hline\hline
%
95\,\% percentile of 10\,\% spectral roll-off point  &      &  \checkmark  & \checkmark   &     &    \\
Mean-crossing rate of 10\,\% spectral roll-off point           &      &    &    &  \checkmark   & \checkmark   \\
3rd quartile of spectral centroid     &      &    &    &  \checkmark   &  \checkmark  \\
\hline 
%
Mean of non-zero values of the spectral flux           &      &  \checkmark  &  \checkmark  &     &    \\
1st quartile of the spectral flux        &      &  \checkmark  &  \checkmark  &     &    \\
Inter-quartile range 3-1 of the spectral flux        &      &    & \checkmark   &     &  \checkmark  \\
\hline\hline
%
3rd quadratic regression coefficient (offset c) of $\Delta$ $F_0$               &      &    &    &  \checkmark   & \checkmark   \\
Skewness of $\Delta\Delta$ $F_0$  &      &    &    &  \checkmark   & \checkmark   \\
Zero-crossing rate of prob.\ of voicing        &      &    &    &  \checkmark   &  \checkmark  \\
Quadratic error of linear regression of $\Delta$ prob.\ of voicing      &      &    &    &  \checkmark   &  \checkmark  \\
\hline\hline
%
(Minimum - mean) of $\Delta$ logarithmic energy        & \checkmark &  \checkmark  &    &     &    \\
\hline\hline
%
Number of peaks of MFCC$_2$           & \checkmark &  \checkmark  &    &     &    \\
Zero-crossing rate of $\Delta$ MFCC$_2$                &      &  \checkmark  &    &     & \checkmark   \\
Inter-quartile range 3-2 of $\Delta$ MFCC$_2$          & \checkmark &  \checkmark  &    &     &    \\
DCT$_0$ of $\Delta\Delta$ MFCC$_2$        & \checkmark &  \checkmark  &    &     &    \\
\hline
Slope (m) of lin.\ approx.\ of MFCC$_4$            & \checkmark &  \checkmark  & \checkmark   &     &    \\
1st quadratic regression coefficient (a) of   MFCC$_4$          & \checkmark &  \checkmark  &  \checkmark  &     &    \\
95\,\% percentile of $\Delta$ MFCC$_4$        &      &    &    &  \checkmark   &  \checkmark  \\
\hline
%
Percentage of falling $\Delta$ MFCC$_5$          & \checkmark &  \checkmark  &    &     &    \\
\hline
Number of peaks of MFCC$_6$          &      &    &    &  \checkmark   &  \checkmark  \\
\hline
Percentage of rising MFCC$_8$                & \checkmark &  \checkmark  &    &     &    \\
DCT$_0$ of $\Delta\Delta$ MFCC$_8$               & \checkmark &  \checkmark  &    &     &    \\
\hline
Minimum of MFCC$_9$            & \checkmark &  \checkmark  & \checkmark   &     &    \\
\hline
(Minimum - mean) of $\Delta$ MFCC$_10$       & \checkmark &  \checkmark  &    &     &    \\
\hline
Zero-crossing rate of $\Delta\Delta$ MFCC$_11$        & \checkmark &  \checkmark  &    &     &    \\
\hline
\hline\hline
%
Position of maximum of minimal raw sample value              & \checkmark &  \checkmark  &  \checkmark  &     &    \\
%
Mean-crossing rate of maximum raw sample value               & \checkmark &  \checkmark  &    &     &    \\
\hline
Position of maximum of $\Delta$ zero-crossing rate           & \checkmark &  \checkmark  &    &     &    \\
3rd quartile of zero-crossing rate            &      &    &    &  \checkmark   & \checkmark   \\
\hline\hline
\end{tabular}
\end{table}

It is not easy to interpret the single features chosen in Tab.\ \ref{tab:commonfeatures}; moreover, it might be that a feature has been preferred by the CFS to another related one due to some spurious factors given in the rather small sample.  
A better way of representing the results is the summary given in Tab.\ \ref{tab:commonfeaturesGroup} where
an overview of predominant acoustic \textit{low-level descriptor categories} and corresponding \textit{functional categories} in the sets are given. Acoustic low-level descriptor categories are used as in Tab.~\ref{tab:lld}: time signal features, energy, voice, spectral, and cepstral. The functionals from Tab.~\ref{tab:functionals} are combined into four categories corresponding to certain physical signal properties: functionals primarily describing the low-level feature contour \textit{Modulation} (DCT
(Discrete Cosine Transform)
 coefficients, zero-/mean-crossing rates, kurtosis, number of peaks, regression error etc.), value \textit{distribution} (max.\ and min.\, ranges, means, percentiles, etc.), relative \textit{position} within the word/turn (relative position of peaks, min.\ value, max.\ value, centroid, etc.), and regression features describing the overall \textit{shape} of the low-level feature contour.

\begin{table}[th]
  \caption{Common features by low-level descriptor and functional group selected in all 51 folds of at least one experiment for turn and word-unit based detection/classification. 2, 3, and 6 classes for word-unit based classification (W$_2$, W$_3$, W$_6$), and 2 and 3 classes for turn based detection (T$_2$, T$_3$).}
    \label{tab:commonfeaturesGroup}
\vspace{0.3cm}
\centering
  \begin{tabular}{ll|l|ccc|cc}  \hline\hline
\multicolumn{3}{c|}{\textbf{Feature description} }           & W$_2$ & W$_3$ & W$_6$    & T$_2$ & T$_3$ \\ \hline
Feature Group & Details & Functionals       &  & & & & \\\hline\hline
Time     & Max.\/Min.\ sample val.\  & modulation & \checkmark & \checkmark & & & \\ 
         &                           & position & \checkmark & \checkmark & \checkmark & & \\ 
         & Zero-crossing rate  & position & \checkmark & \checkmark & & & \\ 
         &                     & distribution &  &  & & \checkmark& \checkmark \\ 
\hline
Energy   & Change of energy          & distribution & \checkmark &\checkmark & & & \\
         &                           & modulation & \checkmark & & & & \\
         & Energy                    & modulation &  & & & \checkmark & \\ 
         &                           & position &  & & &  &\checkmark \\ 
\hline
Pitch    & Change of Change of $F_0$ & distribution & \checkmark & & & & \\
         & Change of $F_0$           & shape, distribution &   &  & &  & \checkmark \\
         & $F_0$                     & shape, position &  & & & \checkmark & \\
         & Voicing Probability       & modulation   &   & \checkmark & & \checkmark & \checkmark \\
         &                           & shape   &   &  & &  & \checkmark \\
         &                           & distribution &   &  & \checkmark & & \\
\hline
Spectral & Energy in voice $F_0$ band   & modulation & \checkmark & & & & \\
         & Frequency distribution          & distribution &  & \checkmark & \checkmark & \checkmark & \checkmark \\
         & Frequency distribution          & modulation &  & \checkmark & & \checkmark & \checkmark \\
         & Flux                & distribution  &  &  &  \checkmark& & \checkmark\\
         & Flux                & modulation  &  &  &  & \checkmark& \checkmark\\
\hline
Cepstral & MFCC 2                    & distribution, modulation & \checkmark & \checkmark& & & \\
         & MFCC 3                    & shape, modulation &  & & &\checkmark & \\
         & MFCC 4                    & distribution & \checkmark & \checkmark & & \checkmark &\checkmark \\
         &                           & modulation & \checkmark & \checkmark & & \checkmark & \\
         &                           & shape & \checkmark & \checkmark & \checkmark & & \checkmark\\
         & MFCC 6                    & distribution, modulation &  & & &\checkmark & \\
         & MFCC 8                    & distribution, modulation & \checkmark & \checkmark & & & \\
         & MFCC 9                    & distribution & \checkmark & \checkmark & \checkmark & & \\ 
\hline\hline
\end{tabular}
\end{table}

The relevant feature groups in Tab.\ \ref{tab:commonfeaturesGroup} can be roughly put into three categories: the first category containing features that are highly relevant for word-unit based as well as turn-based laughter classification/detection, the second category containing features only relevant for word-unit based laughter classification, and the third containing features only relevant for turn-based laughter detection.
Features in the first category are mostly modulation and value distribution statistics of the fourth Mel-frequency cepstral coefficient, the probability of voicing, and parameters describing the distribution of the signal energy among spectral bands (\ie spectral roll-off points, energies in selected frequency bands, and the centroid of the local spectrum). The fact that functionals describing signal modulation are selected often reveals that laughter is characterised by modulation in certain features of the speech signal, which is expected, since laughter has periodically re-occurring elements~\cite{Campbell05-NLM}. The distribution of signal values also describes the quality of the signal modulation. For signals with narrow peaks the mean value will be closer to the minimum value than to the maximum value, whereas for signals with broader peaks the mean value will move up closer towards the maximum value. The associated low-level descriptors in conjunction with the functionals describing modulation reveal that modulation of the spectral distribution and voicing probability can characterise laughter. Generally speaking, these features are describing a pattern of repeating change between voiced and unvoiced segments and associated changes in speech spectra. However, since both the probability of voicing \textit{and} the spectral distribution are relevant, we can conclude that the spectral distribution might be used to discriminate voiced laughter segments from voiced speech segments due to different pitch and formant characteristics. 
For word-based classification only (second category), next to distribution and modulation functionals, position functionals (\ie the relative positions of minima or maxima within the word-unit) play an important role. Considering that mostly energy and amplitude related low-level descriptors are associated with these position features, this might be linked to word prosody. A word has prominence on one specific syllable -- at least this is the case for German word accent position -- while laughter has multiple, periodically occurring segments which are 'emphasised'. 
For turn-based detection only (third category), the modulation of the spectral flux (\ie change in the spectrum over time) is shown to be important. This indicates -- considering the turn-level detection task -- that as soon as laughter is present in the analysed turn, the overall modulation of spectral change seems to significantly differ from regular speech. 

\section{Concluding Remarks}
\label{Concluding Remarks}

Notable is the relatively small number of laughter instances in our relatively large database: only 0.4\,\%. This is some disadvantage because  of sparse data -- especially considering the fact that speakers employ laughter in different ways. On the other hand, we can consider it an advantage as well being able to investigate realistic data where laughter neither was elicited nor facilitated via selection of speakers or tasks. The distribution of the types of laughter we have found indicates a highly developed system with specific functions and positions of laughter. We found strong tendencies, \eg for \textsc{speech-laugh} not to occur internally inside syntactic units; the same tendency had no exception for \textsc{laughter} -- we can call it a rule, probably with no exceptions. 
The `punctuation' function is weaker but still visible in the dialogue (task) structure.
The feature groups surviving our correlation-based feature-subset selection (CFS) procedure give a clear picture of the acoustic characteristics of laughter. Classifying laughter automatically is a difficult task; this holds especially for telling apart different types belonging to the same main class. 
On the other hand, detecting \textbf{some} laughter in a turn seems to be promising for foreseeable applications, especially if we do not aim at single instance detection but at a summarizing estimation of a general degree of `laughter proneness' in an interaction.

Our results show that children -- at least at the age of 10--13 years --  fully master the interplay of non-verbal/paralinguistic events such as laughter with syntactic structure and dialogue structure. Also the communicative functions of laughter  seem not to be different from the use of laughter known so far from studies with adult human-human interactions. 

\section{Acknowledgments}
\label{Acknowledgments}
The research leading to these results has received funding from the
European Community
under grant No. IST-2001-37599 (PF-STAR),
grant No. IST-2002-50742 (HUMAINE), 
and grant (FP7/2007-2013) No. 211486 (SEMAINE).
The responsibility lies with the authors.

\vspace{1cm}

\textbf{Note:} This manuscript should have been part of a book with the title `Phonetics of Laughter', edited by J. Trouvian and N. Campbell, targeted for publication 2011-2014; however, this book never appeared. References to pertinent literature are not updated.

\newpage

\bibliographystyle{IEEEtran}
\bibliography{batliner-emo}

\begin{thebibliography}{10}
\providecommand{\url}[1]{#1}
\csname url@rmstyle\endcsname
\providecommand{\newblock}{\relax}
\providecommand{\bibinfo}[2]{#2}
\providecommand\BIBentrySTDinterwordspacing{\spaceskip=0pt\relax}
\providecommand\BIBentryALTinterwordstretchfactor{4}
\providecommand\BIBentryALTinterwordspacing{\spaceskip=\fontdimen2\font plus
\BIBentryALTinterwordstretchfactor\fontdimen3\font minus
  \fontdimen4\font\relax}
\providecommand\BIBforeignlanguage[2]{{%
\expandafter\ifx\csname l@#1\endcsname\relax
\typeout{** WARNING: IEEEtran.bst: No hyphenation pattern has been}%
\typeout{** loaded for the language `#1'. Using the pattern for}%
\typeout{** the default language instead.}%
\else
\language=\csname l@#1\endcsname
\fi
#2}}

\bibitem{Batliner07-LAE}
A.~Batliner, S.~Steidl, and E.~N{\"o}th, ``{Laryngealizations and Emotions: How
  Many Babushkas?}'' in \emph{Proceedings of the International Workshop on
  Paralinguistic Speech -- between Models and Data (ParaLing'07)},
  Saarbr{\"u}cken, 2007, pp. 17--22.

\bibitem{Schroeder00-ESO}
M.~Schr{\"o}der, ``Experimental study of affect bursts,'' in \emph{Proceedings
  of the ISCA Workshop on Speech and Emotion}, Newcastle, Northern Ireland,
  2000, pp. 132--137.

\bibitem{Darwin72-TEO}
C.~Darwin, \emph{The Expression of the Emotions in Man and Animals}.\hskip 1em
  plus 0.5em minus 0.4em\relax London: John Murray, 1872, (P.~Ekman, Ed.,
  Oxford University Press, Oxford, 3. Ed., 1998).

\bibitem{Provine93-LPS}
R.~Provine, ``{Laughter punctuates speech: linguistic, social and gender
  contexts of laughter},'' \emph{Ethology}, vol.~15, pp. 291--298, 1993.

\bibitem{Bacharowski01-TAF}
J.-A. Bacharowski and M.~J. Smoski, ``{The acoustic features of human
  laughter},'' \emph{Journal of the Acoustical Society of America}, vol. 110,
  no.~3, pp. 1581--1597, 2001.

\bibitem{Trouvain01-PAO}
J.~Trouvain, ``{Phonetic Aspects of ``Speech Laughs''},'' in \emph{Proceedings
  of the Conference on Orality and Gestuality Orage 2001}, Aix-en-Provence,
  2001, pp. 634--639.

\bibitem{Trouvain03-SPU}
------, ``{Segmenting Phonetic Units in Laughter},'' in \emph{Proc. ICPhS},
  Barcelona, 2003, pp. 2793--2796.

\bibitem{Campbell05-NLM}
N.~Campbell, H.~Kashioka, and R.~Ohara, ``No laughing matter,'' in \emph{Proc.\
  Interspeech}, Lisbon, 2005, pp. 465--468.

\bibitem{Campbell07-WWL}
N.~Campbell, ``{Whom we laugh with affects how we laugh},'' in
  \emph{Proceedings of the Interdisciplinary Workshop on The Phonetics of
  Laughter}, J.~Trouvain and N.~Campbell, Eds., Saarbr{\"u}cken, 2007, pp.
  61--65.

\bibitem{Laskowski07-AOT}
K.~Laskowski and S.~Burger, ``Analysis of the occurence of laughter in
  meetings,'' in \emph{Proc.\ Interspeech}, Antwerp, 2007, pp. 1258--1261.

\bibitem{Truong05-ADO}
K.~Truong and D.~van Leeuwen, ``Automatic detection of laughter,'' in
  \emph{Proc.\ Interspeech}, Lisbon, Portugal, 2005, pp. 485--488.

\bibitem{Truong07-ADB}
K.~P. Truong and D.~A. van Leeuwen, ``{Automatic discrimination between
  laughter and speech},'' \emph{Speech Communication}, vol.~49, pp. 144--158,
  2007.

\bibitem{Petridis08-ALD}
S.~Petridis and M.~Pantic, ``{Audiovisual Laughter Detection based on Temporal
  Features},'' in \emph{Proc. ICMI'08}, Chania, 2008, pp. 37--44.

\bibitem{Batliner08-PEV}
A.~Batliner, S.~Steidl, C.~Hacker, and E.~N{\"o}th, ``{Private emotions vs.
  social interaction --- a data-driven approach towards analysing emotions in
  speech},'' \emph{User Modeling and User-Adapted Interaction}, vol.~18, pp.
  175--206, 2008.

\bibitem{Steidl09-ACO}
S.~Steidl, \emph{Automatic Classification of Emotion-Related User States in
  Spontaneous Children's Speech}.\hskip 1em plus 0.5em minus 0.4em\relax
  Berlin: Logos Verlag, 2009, (PhD thesis, FAU Erlangen-Nuremberg).

\bibitem{Schuller09-TI2}
B.~Schuller, S.~Steidl, and A.~Batliner, ``{The INTERSPEECH 2009 Emotion
  Challenge},'' in \emph{Proc.\ Interspeech}, Brighton, 2009, pp. 312--315.

\bibitem{Ekman99-BE}
P.~Ekman, ``Basic emotions,'' in \emph{Handbook of Cognition and Emotion},
  T.~Dalgleish and M.~Power, Eds.\hskip 1em plus 0.5em minus 0.4em\relax New
  York: John Wiley, 1999, pp. 301--320.

\bibitem{Ortony88-TCS}
A.~Ortony, G.~L. Clore, and A.~Collins, \emph{The Cognitive Structure of
  Emotions}.\hskip 1em plus 0.5em minus 0.4em\relax Cambridge, New York:
  Cambridge University Press, 1988.

\bibitem{Batliner05-TOT}
A.~Batliner, S.~Steidl, C.~Hacker, E.~N{\"o}th, and H.~Niemann, ``{Tales of
  Tuning -- Prototyping for Automatic Classification of Emotional User
  States},'' in \emph{{Proc.\ Interspeech}}, {Lisbon}, 2005, pp. 489--492.

\bibitem{Kennedy04-LDI}
L.~Kennedy and D.~Ellis, ``{Laughter detection in meetings},'' in \emph{Proc.
  ICASSP Meeting Recognition Workshop}, Montreal, 2004, pp. 118--121.

\bibitem{Laskowski08-DOL}
K.~Laskowski and T.~Schultz, ``Detection of laughter-in-interaction in
  multichannel close-talk microphone recordings of meetings,'' in
  \emph{{Machine Learning for Multimodal Interaction}}, ser. {Lecture Notes in
  Computer Science 5237}, A.~Popescu-Belis and R.~Stiefelhagen, Eds.,
  Berlin-Heidelberg, 2008, pp. 149--160.

\bibitem{Keyton10-ELF}
J.~Keyton and S.~J. Beck, ``{Examining Laughter Functionality in Jury
  Deliberations},'' \emph{Small Group Research}, vol.~41, pp. 386--407, 2010.

\bibitem{Batliner98-MSP}
A.~Batliner, R.~Kompe, A.~Kie{\ss}ling, M.~Mast, H.~Niemann, and E.~N{\"o}th,
  ``{M $=$ Syntax $+$ Prosody: A syntactic--prosodic labelling scheme for large
  spontaneous speech databases},'' \emph{Speech Communication}, vol.~25, no.~4,
  pp. 193--222, September 1998.

\bibitem{Eysenck60-TCO}
H.~Eysenck, ``{The Concept of Statistcal Significance and the Controversy about
  One-Tailed Tests},'' \emph{Psychological Review}, vol.~67, pp. 269--271,
  1960.

\bibitem{Rozeboom60-TFO}
W.~Rozeboom, ``{The Fallacy of the Null-Hypothesis Significance Test},''
  \emph{Psychological bulletin}, vol.~57, pp. 416--428, 1960.

\bibitem{Garrett66-TAU}
M.~Garrett, T.~Bever, and J.~Fodor, ``{The active use of grammar in speech
  perception},'' \emph{Perception and Psychophysics}, vol.~1, pp. 30--32, 1966.

\bibitem{Nwokah99-TIO}
E.~E. Nwokah, H.-C. Hsu, and P.~Davies, ``{The Integration of Laughter and
  Speech in Vocal Communication},'' \emph{Journal of Speech, Language, and
  Hearing Research}, vol.~42, pp. 880--894, 1999.

\bibitem{Batliner08-TTO}
A.~Batliner, C.~Hacker, and E.~N{\"o}th, ``{To Talk or not to Talk with a
  Computer -- Taking into Account the User's Focus of Attention},''
  \emph{Journal on Multimodal User Interfaces}, vol.~2, pp. 171--186, 2008.

\bibitem{Batliner06-TPO}
A.~Batliner, S.~Biersack, and S.~Steidl, ``{The Prosody of Pet Robot Directed
  Speech: Evidence from Children},'' in \emph{{Proceedings of Speech Prosody
  2006}}, {Dresden}, 2006, pp. 1--4.

\bibitem{Batliner08-MAC}
A.~Batliner, B.~Schuller, S.~Schaeffler, and S.~Steidl, ``{Mothers, Adults,
  Children, Pets --- Towards the Acoustics of Intimacy},'' in \emph{{Proc.\
  ICASSP 2008}}, Las Vegas, 2008, pp. 4497--4500.

\bibitem{Eyben09-OIT}
F.~Eyben, M.~W{\"o}llmer, and B.~Schuller, ``{openEAR - Introducing the Munich
  Open-Source Emotion and Affect Recognition Toolkit},'' in \emph{Proc.\ ACII},
  Amsterdam, 2009, pp. 576--581.

\bibitem{Schuller06-MST}
B.~Schuller, F.~Wallhoff, D.~Arsic, and G.~Rigoll, ``Musical signal type
  discrimination based on large open feature sets,'' in \emph{Proceedings of
  the International Conference on Multimedia {\&} Expo ICME 2006}.\hskip 1em
  plus 0.5em minus 0.4em\relax IEEE, 2006.

\bibitem{Grimm07-AER}
M.~Grimm, K.~Kroschel, B.~Schuller, G.~Rigoll, and T.~Moosmayr, ``Acoustic
  emotion recognition in car environment using a 3d emotion space approach,''
  in \emph{Proceedings of the DAGA 2007}.\hskip 1em plus 0.5em minus
  0.4em\relax Stuttgart, Germany: DEGA, March 2007, pp. 313--314.

\bibitem{Schuller07-TRO}
B.~Schuller, A.~Batliner, D.~Seppi, S.~Steidl, T.~Vogt, J.~Wagner,
  L.~Devillers, L.~Vidrascu, N.~Amir, L.~Kessous, and V.~Aharonson, ``{The
  Relevance of Feature Type for the Automatic Classification of Emotional User
  States: Low Level Descriptors and Functionals},'' in \emph{Proc.\
  Interspeech}, Antwerp, 2007, pp. 2253--2256.

\bibitem{Schuller08-SAD}
B.~Schuller, F.~Eyben, and G.~Rigoll, ``Static and dynamic modelling for the
  recognition of non-verbal vocalisations in conversational speech,'' in
  \emph{Proceedings of the 4th IEEE Tutorial and Research Workshop on
  Perception and Interactive Technologies for Speech-based Systems (PIT 2008),
  Kloster Irsee, Germany}, E.~Andr{\'e}, Ed., vol. LNCS 5078.\hskip 1em plus
  0.5em minus 0.4em\relax Springer, 2008, pp. 99--110.

\bibitem{Witten05-DMP}
I.~H. Witten and E.~Frank, \emph{{Data mining: Practical machine learning tools
  and techniques, 2nd Edition}}.\hskip 1em plus 0.5em minus 0.4em\relax San
  Francisco: Morgan Kaufmann, 2005.

\bibitem{Vlasenko07-FVT}
B.~Vlasenko, B.~Schuller, A.~Wendemuth, and G.~Rigoll, ``{Frame vs. Turn-Level:
  Emotion Recognition from Speech Considering Static and Dynamic Processing},''
  in \emph{Affective Computing and Intelligent Interaction}, A.~Paiva,
  R.~Prada, and R.~W. Picard, Eds.\hskip 1em plus 0.5em minus 0.4em\relax
  Berlin-Heidelberg: Springer, 2007, pp. 139--147.

\end{thebibliography}

\end{document}